\documentclass[11pt,letterpaper]{article}
\usepackage[letterpaper]{geometry}
\usepackage{acl2012}
\usepackage{times}
\usepackage{latexsym}
\usepackage{amsmath}
\usepackage{graphicx}
\usepackage{multirow}
\usepackage{arydshln}
\usepackage{url}
\usepackage{subcaption}
\usepackage[utf8]{inputenc}
\usepackage{lipsum}
\newcommand{\tabincell}[2]{\begin{tabular}{@{}#1@{}}#2\end{tabular}}

\makeatletter
\newcommand{\@BIBLABEL}{\@emptybiblabel}
\newcommand{\@emptybiblabel}[1]{}
\makeatother
\usepackage[hidelinks]{hyperref}

\title{Adversarial Neural Machine Translation}

\author{Lijun Wu$^1$, Yingce Xia$^2$, Li Zhao$^3$, Fei Tian$^3$, Tao Qin$^3$, Jianhuang Lai$^{1,4}$ \and Tie-Yan Liu$^3$\\
$^1$School of Data and Computer Science, Sun Yat-sen University\\ \quad$^2$University of Science and Technology of China \\
\quad$^3$Microsoft Research Asia \\ \quad$^4$Guangdong Key Laboratory of Information Security Technology\\
wulijun3@mail2.sysu.edu.cn;\;yingce.xia@gmail.com;\;\\
\{lizo,fetia,taoqin,tie-yan.liu\}@microsoft.com;\;stsljh@mail.sysu.edu.cn
}

\date{}

\begin{document}
\maketitle
\begin{abstract}
In this paper, we study a new learning paradigm for Neural Machine Translation (NMT). Instead of maximizing the likelihood of the human translation as in previous works, we minimize the distinction between human translation and the translation given by an NMT model. To achieve this goal, inspired by the recent success of Generative Adversarial Networks (GANs), we employ an adversarial training architecture and name it as Adversarial-NMT. In Adversarial-NMT, the training of the NMT model is assisted by an adversary, which is an elaborately designed Convolutional Neural Network (CNN). The goal of the adversary is to differentiate the translation result generated by the NMT model from that by human. The goal of the NMT model is to produce high quality translations so as to cheat the adversary. A policy gradient method is leveraged to co-train the NMT model and the adversary. Experimental results on English$\rightarrow$French and German$\rightarrow$English translation tasks show that Adversarial-NMT can achieve significantly better translation quality than several strong baselines.
\end{abstract}

\section{Introduction}
Neural Machine Translation (NMT)~\cite{EncDec,NMT} has drawn more and more attention in both academia and industry~\cite{OpenNMT,NMTLarge,RL4NMT,Coverage4NMT,NMTMolingual,GoogleNMT}. Compared with traditional Statistical Machine Translation (SMT)~\cite{SMT2}, NMT achieves similar or even better translation results in an end-to-end framework. The sentence level maximum likelihood principle and gating units in LSTM/GRU~\cite{LSTM,EncDec}, together with attention mechanisms grant NMT with the ability to better translate long sentences.

Despite its success, the translation quality of latest NMT systems is still far from satisfaction and there remains large room for improvement. For example, NMT usually adopts the Maximum Likelihood Estimation (MLE) principle for training, i.e., to maximize the probability of the target ground-truth sentence conditioned on the source sentence. Such an objective does not guarantee the translation results to be natural, sufficient, and accurate compared with ground-truth translation by human. There are previous works~ \cite{PG4Sequence,RL4NMT,AC4SequencePrediction} that aim to alleviate such limitations of maximum likelihood training, by adopting sequence level objectives (e.g., directly maximizing BLEU~\cite{BLEU}), to reduce the objective inconsistency between NMT training and inference. Yet somewhat improved, such objectives still cannot fully bridge the gap between NMT translations and ground-truth translations.

We, in this paper, adopt a thoroughly different training objective for NMT, targeting at directly minimizing the difference between human translation and the translation given by an NMT model. To achieve this target, inspired by the recent success of Generative Adversarial Networks (GANs)~\cite{GAN}, we design an adversarial training protocol for NMT and name it as Adversarial-NMT. In Adversarial-NMT, besides the typical NMT model, an adversary is introduced to distinguish the translation generated by NMT from that by human (i.e., ground truth). Meanwhile the NMT model tries to improve its translation results such that it can successfully \emph{cheat} the adversary.

These two modules in Adversarial-NMT are co-trained, and their performances get mutually improved. In particular, the discriminative power of the adversary can be improved by learning from more and more training samples (both positive ones generated by human and negative ones sampled from NMT); and the ability of the NMT model in \emph{cheating} the adversary can be improved by taking the output of the adversary as reward. In this way, the NMT translation results are \emph{professor forced}~\cite{professorForcing} to be as close as possible to ground-truth translation.

Different from previous GANs, which assume the existence of a generator in continuous space, in our proposed framework, the NMT model is in fact not a typical generative model, but instead a probabilistic transformation that maps a source language sentence to a target language sentence, both in discrete space. Such differences make it necessary to design both new network architectures and optimization methods to make adversarial training possible for NMT. We therefore on one aspect, leverage a specially designed Convolutional Neural Network (CNN) model as the adversary, which takes the (source, target) sentence pair as input; on the other aspect, we turn to a policy gradient method named REINFORCE~\cite{REINFORCE}, widely used in the reinforcement learning literature~\cite{RLSutton}, to guarantee both the two modules are effectively optimized in an adversarial manner. We conduct extensive experiments, which demonstrates that Adversarial-NMT can achieve significantly better translation results than traditional NMT models with even much larger vocabulary size and higher model complexity.

\section{Related Work}
\label{sec:related}
End-to-end Neural Machine Translation (NMT)~\cite{NMT,EncDec,S2S,NMTLarge,GoogleNMT,BaiduNMT} has been the recent research focus of the community. Typical NMT system is built within the RNN based encoder-decoder framework. In such a framework the encoder RNN sequentially processes the words in a source language sentence into fixed length vectors, which act as the inputs to decoder RNN to decode the translation sentence. NMT typically adopts the principle of Maximum Likelihood Estimation (MLE) for training, i.e., maximizing the per-word likelihood of target sentence. Other training criteria, such as Minimum Risk Training (MRT) based on reinforcement learning~\cite{PG4Sequence,RL4NMT} and translation reconstruction~\cite{NMTReconstruction}, are shown to improve over such word-level MLE principle since these objectives take the translation sentence as a whole.

The training principle we propose is based on the spirit of Generative Adversarial Networks (GANs)~\cite{GAN,ImprovedGAN}, or more generally, adversarial training~\cite{AdvExamples}. In adversarial training, a discriminator and a generator compete with each other, forcing the generator to produce high quality outputs that are able to fool the discriminator. Adversarial training typically succeed in image generation~\cite{GAN,GANTOI}, with limited contribution in natural language processing tasks~\cite{seqGAN,GAN4Bot}, mainly due to the difficulty of propagating the error signals from the discriminator to the generator through the discretely generated natural language tokens.~\newcite{seqGAN} alleviates such a difficulty by reinforcement learning approach for sequence (e.g., music) generation. However, as far as we know, there are limited efforts on adversarial training for sequence-to-sequence task when a conditional mapping between two sequences is involved, and our work is among the first endeavors to explore the potential of acting in this way, especially for Neural Machine Translation \cite{YangGAN}.

\section{Adversarial-NMT}
\label{sec:adv-nmt}

\begin{figure}[!htpb]
\centering
\includegraphics[width=1.05\linewidth]{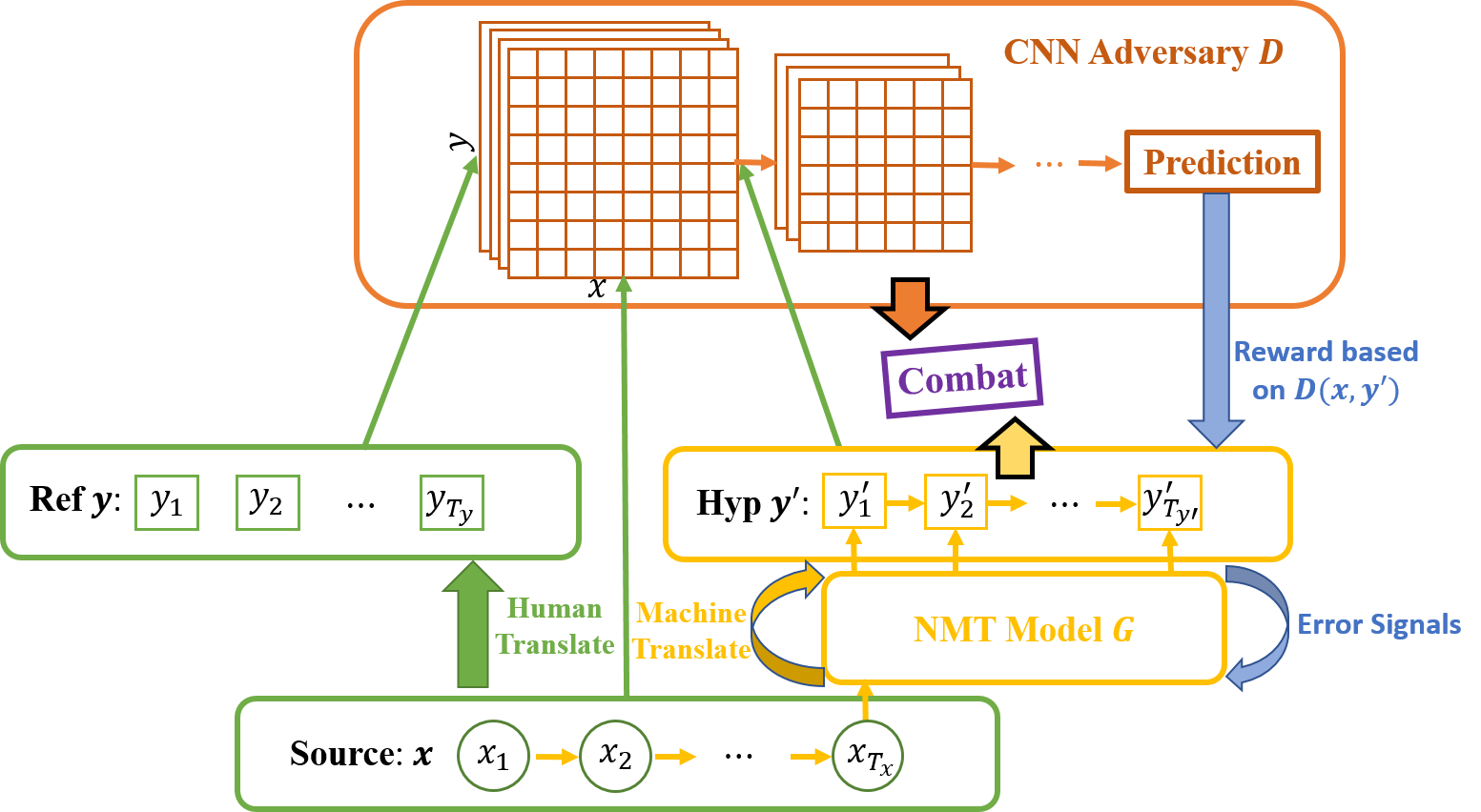}
\caption{The Adversarial-NMT framework.  `Ref' is short for `Reference' which means the ground-truth translation and `Hyp' is short for `Hypothesis', denoting model translation sentence. All the yellow parts denote the NMT model $G$, which maps a source sentence $x$ to a translation sentence. The red parts are the adversary network $D$, which predicts whether a given target sentence is the ground-truth translation of the given source sentence $x$. $G$ and $D$ combat with each other, generating both sampled translation $y'$ to train $D$, and the reward signals to train $G$ by policy gradient (the blue arrows).}
\label{fig:framework}
\end{figure}

The overall framework of Our Adversarial-NMT is shown in Figure \ref{fig:framework}. Let $(x=\{x_1,x_2,...,x_{T_x}\} , y=\{y_1,y_2,...,y_{T_y}\})$ be a bilingual aligned sentence pair for training, where $x_i$ is the $i$-th word in the source sentence and $y_j$ is the $j$-th word in the target sentence. Let $y'$ denote the translation sentence out from an NMT system for the source sentence $x$. As previously stated, the goal of Adversarial-NMT is to force $y'$ to be as `similar' as $y$. In the perfect case, $y'$ is so similar to the human translation $y$ that even a human cannot tell whether $y'$ is generated by machine or human. In order to achieve that, we introduce an extra adversary network, which acts similarly to the discriminator adopted in GANs~\cite{GAN}. The goal of the adversary is to differentiate human translation from machine translation, and the NMT  model $G$ tries to produce a target sentence as similar as human translation so as to fool the adversary.

\subsection{NMT Model}
We adopt the Recurrent Neural Network (RNN) based encoder-decoder as the NMT model to seek a target language translation $y'$ given source sentence $x$. In particular, a probabilistic mapping $G(y|x)$ is firstly learnt and the translation result $y'\sim G(\cdot|x)$ is sampled from it. To be specific, given source sentence $x$ and previously generated words $y_{<t}$, the probability of generating word $y_t$ is:
\begin{eqnarray}
G(y_t|y_{<t},x) \propto \exp(y_t; r_t, c_t)\\
r_{t}=g(r_{t-1},y_{t-1},c_t)
\end{eqnarray}
where $r_t$ is the decoding state from decoder at time $t$. Here $g$ is the recurrent unit such as the Long Short Term Memory (LSTM) unit \cite{LSTM} or Gated Recurrent Unit (GRU) \cite{EncDec}, and $c_t$ is a distinct source representation at time $t$ calculated by an attention mechanism \cite{NMT}:
\begin{eqnarray}
  c_t=\sum_{i=1}^{T_x}\alpha_{it}h_i \\
  \alpha_{it} = \frac{\exp\{a(h_i, r_{t-1})\}}{\sum_j\exp\{a(h_j,r_{t-1})\}}
\end{eqnarray}
where $T_x$ is the source sentence length, $a(\cdot , \cdot)$ is a feed-forward neural network and $h_i$ is the hidden state from RNN encoder computed by $h_{i-1}$ and $x_i$:
\begin{eqnarray}
h_i=f(h_{i-1},x_i)
\end{eqnarray}
The translation result $y'$ can be sampled from $G(\cdot|x)$ either in a greedy way for each timestep, or using beam search~\cite{S2S} to seek globally optimized result.

\subsection{Adversary Model}
The adversary is used to differentiate translation result $y'$ and the ground-truth translation $y$, given the source language sentence $x$. To achieve that, one needs to measure the translative matching degree of source-target sentence pair $(x,y)$. We turn to Convolution Neural Network (CNN) for this task~\cite{ABCNN,MatchingCNN}, since with its layer-by-layer convolution and pooling strategies, CNN is able to accurately capture the hierarchical correspondence of $(x,y)$ at different abstraction levels.

\begin{figure}[!htb]
\centering
\includegraphics[width=1.05\linewidth]{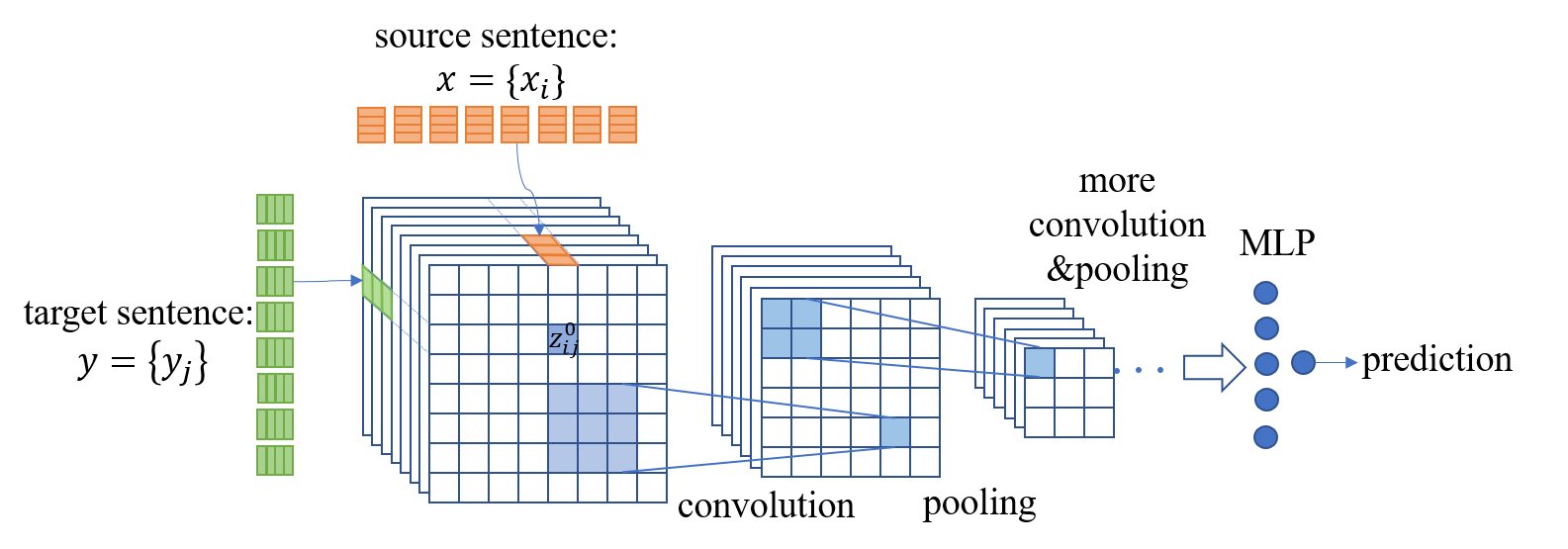}
\caption{The CNN adversary framework.}
\label{fig:cnn_framework}
\end{figure}

The general structure is shown in Figure \ref{fig:cnn_framework}. Specifically, given a sentence pair $(x,y)$, we first construct a $2D$ image-like representation by simply concatenating the embedding vectors of words in $x$ and $y$. That is, for $i$-th word $x_i$ in $x$ and $j$-th word $y_j$ in sentence $y$, we have the following feature map:

\begin{equation}
z_{i,j}^{(0)} = [x_{i}^{T}, y_{j}^{T}]^{T} \nonumber
\end{equation}

Based on such a $2D$ image-like representation, we perform convolution on every $3\times 3$ window, with the purpose to capture the correspondence between segments in $x$ and segments in $y$ by the following feature map of type $f$:

\begin{equation}
z_{i,j}^{(1, f)} = \sigma(W^{(1,f)}z_{i,j}^{(0)} + b^{(1,f)}) \nonumber
\end{equation}
where $\sigma(\cdot)$ is the sigmoid active function, $\sigma(x) = 1/(1 + \exp(-x))$.

After that we perform a max-pooling in non-overlapping $2\times2$ windows:

\begin{equation}
z_{i,j}^{(2, f)}\!=\!\max(\{z_{2i-1,2j-1}^{(1, f)}, z_{2i-1,2j}^{(1, f)}, z_{2i,2j-1}^{(1, f)}, z_{2i,2j}^{(1, f)}\}) \nonumber
\end{equation}

We could go on for more layers of convolution and max-pooling, aiming at capturing the correspondence at different levels of abstraction. The extracted features are then fed into a multi-layer perceptron (MLP), with sigmoid activation at the last layer to give the probability that $(x,y)$ is from ground-truth data, i.e. $D(x,y)$. The optimization target of such CNN adversary is to minimize the cross entropy loss for binary classification, with ground-truth data $(x,y)$ as positive instance while sampled data (from $G$) $(x,y')$ as negative one.


\subsection{Policy Gradient Algorithm to Train Adversarial-NMT}
With the notations for NMT model $G$ and adversary model $D$, the final training objective is:

\begin{equation}
\label{eq:minimaxgame-definition}
\begin{aligned}
 &\min_G \max_D V(D, G)\\
 =&E_{(x,y)\sim P_{data}(x,y)}[\log D(x,y)] + \\
 &E_{x\sim P_{data}(x), y'\sim G(\cdot|x)}[\log (1-D(x,y'))]
\end{aligned}
\end{equation}

That is, translation model $G$ tries to produce high quality translation $y'$ to fool the adversary $D$ (the outer-loop $\min$), whose objective is to successfully classify translation results from real data (i.e., ground-truth) and from $G$ (the inner-loop $\max$).

Eqn.~\eqref{eq:minimaxgame-definition} reveals that it is straightforward to train the adversary $D$, by keeping providing $D$ with the ground-truth sentence pair $(x,y)$ and the sampled translation pair $(x,y')$ from $G$, respectively as positive and negative training data. However, when it turns to NMT model $G$, it is non-trivial to design the training process, given that the discretely sampled $y'$ from $G$ makes it difficult to directly back-propagate the error signals from $D$ to $G$, making $V(D,G)$ nondifferentiable w.r.t. $G$'s model parameters $\Theta_G$.

To tackle the above challenge, we leverage the REINFORCE algorithm ~\cite{REINFORCE}, a Monte-Carlo policy gradient method in reinforcement learning literature to optimize $G$. Note that the objective of training $G$ under a fixed source language sentence $x$ and $D$ is to minimize the following loss item:

\begin{equation}
\label{eqn:obj_G}
L=E_{y'\sim G(\cdot|x)}\log(1-D(x,y'))
\end{equation}
whose gradient w.r.t. $\Theta_G$ is:

\begin{equation}
\label{eqn:grad_G}
\begin{aligned}
&\nabla_{\Theta_G}L \\
=&\nabla_{\Theta_G}E_{y'\sim G(\cdot|x)}[\log(1-D(x,y'))]\\
=&E_{y'\sim G(\cdot|x)}[\log(1-D(x,y'))\nabla_{\Theta_G}\log G(y'|x)]
\end{aligned}
\end{equation}

A sample $y'$ from $G(\cdot|x)$ is used to approximate the above gradient:
\begin{equation}
\label{eqn:grad_G_approx}
\nabla_{\Theta_G}\approx\!\hat\nabla_{\Theta_G}\!=\!\log(1-D(x,y'))\nabla_{\Theta_G}\log G(y'|x)
\end{equation} in which $\nabla_{\Theta_G}\log G(y'|x)$ are gradients specified with standard sequence-to-sequence NMT networks. Such a gradient approximation is used to update $\Theta_G$:

\begin{equation}
\Theta_G\leftarrow\Theta_G -\alpha \hat\nabla_{\Theta_G}
\end{equation}where $\alpha$ is the learning rate.

Using the language of reinforcement learning, in the above Eqn.~\eqref{eqn:obj_G} to \eqref{eqn:grad_G_approx}, the NMT model $G(\cdot|x)$ is the conditional \emph{policy} faced with $x$, while the term $-\log(1-D(x,y'))$, provided by the adversary $D$, acts as a Monte-Carlo estimation of the \emph{reward}. Intuitively speaking, Eqn.~\eqref{eqn:grad_G_approx} implies, the more likely $y'$ to successfully fool $D$ (i.e, larger $D(x,y')$), the larger reward the NMT model will get, and the 'pseudo' training data $(x, y')$ will correspondingly be more favored to improve the policy $G(\cdot|x)$.

Note here we in fact use one sample $-\log(1-D(x,y'))$ from a trajectory $y'$ to estimate the terminal reward given by $D$. Acting in this way brings high variance, to reduce the variance, a moving average of the historical reward values is set as a reward baseline \cite{rewardbaseline}. One can sample multiple trajectories in each decoding step, by regarding $G$ as the roll-out policy to reduce estimation variance for immediate reward~\cite{AlphaGo,seqGAN}.  However, empirically we find such approach is intolerably time-consuming in our task, given that the decoding space in NMT is typically extremely large (the same as vocabulary size).

It is worth comparing our adversarial training with existing methods that directly maximize sequence level measure such as BLEU~\cite{PG4Sequence,RL4NMT,AC4SequencePrediction} in training NMT models, using similar approaches based on reinforcement learning as ours.  We argue that Adversarial-NMT makes the optimization easier compared with these methods. Firstly, the reward learned by our adversary $D$ provides rich and global information to evaluate the translation, which goes beyond the BLEU's simple low-level n-gram matching criteria. Acting in this way provides much smoother objective compared with BLEU since the latter is highly sensitive for slight translation difference at word or phrase level.   Secondly, the NMT model $G$ and the adversary $D$ in Adversarial-NMT co-evolves. The dynamics of adversary $D$ makes NMT model $G$ grows in an adaptive way rather than controlled by a fixed evaluation metric as BLEU. Given the above two reasons, Adversarial-NMT makes the optimization process towards sequence level objectives much more robust and better controlled, which is further verified by its superior performances to the aforementioned methods that will be reported in the next Section~\ref{sec:exp}.

\section{Experiments}
\label{sec:exp}

\subsection{Settings}
We report the experimental results on both English$\rightarrow$French translation (En$\rightarrow$Fr for short) and German$\rightarrow$English translation (De$\rightarrow$En for short).

{\bf Dataset}: For En$\rightarrow$Fr translation, for the sake of fair comparison with previous works, we use the same dataset as~\cite{NMT,RL4NMT}. The dataset is composed of a subset of WMT 2014 training corpus as training set, the combination of news-test 2012 and news-test 2013 as dev set and news-test 2014 as test set, which respectively contains roughly $12M$, $6k$ and $3k$ sentence pairs. The maximal sentence length is 50. We use top $30k$ most frequent English and French words and replace the other words as `UNK' token.

For De$\rightarrow$En translation, following previous works~\cite{PG4Sequence,AC4SequencePrediction}, the dataset is from IWSLT 2014 evaluation campaign~\cite{IWSLT}, consisting of training/dev/test corpus with approximately $153k$, $7k$ and $6.5k$ bilingual sentence pairs respectively. The maximal sentence length is also set as 50.  The dictionary for English and German corpus respectively  include $22,822$ and $32,009$ most frequent words~\cite{AC4SequencePrediction},  with other words replaced as a special token `UNK'.
\begin{table*}[!t]
\small
\centering
\begin{minipage}{14cm}
\begin{tabular}{|l | l | c | c | c}
\hline
System & System Configurations  & BLEU \\
\hline \hline
\multicolumn{4}{c}{{\em Representative end-to-end NMT systems}} \\
\hline
Sutskever et al. \shortcite{S2S} & LSTM with 4 layers + $80K$ vocabs & 30.59 \\
Bahdanau et al. \shortcite{NMT} & RNNSearch & 29.97\footnote{Reported in \protect~\cite{NMTLarge}.} \\
Jean et al. \shortcite{NMTLarge} & RNNSearch + UNK Replace & 33.08 \\
Jean et al. \shortcite{NMTLarge} & RNNSearch + $500k$ vocabs + UNK Replace  & 34.11 \\
Luong et al. \shortcite{RareWordsNMT} & LSTM with 4 layers + $40K$ vocabs & 29.50 \\
Luong et al. \shortcite{RareWordsNMT} & LSTM with 4 layers + $40K$ vocabs + PosUnk & 31.80 \\
Shen et al. \shortcite{RL4NMT} &RNNSearch +Minimum Risk Training Objective& 31.30\\
Sennrich et al. \shortcite{NMTMolingual} & RNNSearch +Monolingual Data & 30.40 \footnote {Reported in \protect~\cite{DualNMT}.} \\
He et al. \shortcite{DualNMT} & RNNSearch+ Monolingual Data + Dual Objective & 32.06\\
\hline
\multicolumn{4}{c}{{\em Adversarial-NMT}} \\
\hline
\multirow{3}{*{\em this work}}{} & RNNSearch + Adversarial Training Objective & 31.91$\dagger$ \\
& RNNSearch + Adversarial Training Objective + UNK Replace & 34.78 \\
\hline
\end{tabular}
\end{minipage}
\caption{Different NMT systems' performances on En$\rightarrow$Fr translation. The default setting is single layer GRU + $30k$ vocabs + MLE training objective, trained with no monolingual data, i.e., the RNNSearch model proposed by Bahdanau et al. \protect\shortcite{NMT}. $\dagger$: significantly better than \protect\newcite{RL4NMT} ($\rho <$ 0.05). } \label{tbl:result_e2f}
\end{table*}

{\bf Implementation Details}: In Adversarial-NMT, the structure of the NMT model $G$ is the same as RNNSearch model~\cite{NMT}, a RNN based encoding-decoding framework with attention mechanism. Single layer GRUs act as encoder and decoder. For En$\rightarrow$Fr translation, the dimensions of word embedding and GRU hidden state are respectively set as $620$ and $1000$, and for De$\rightarrow$En translation they are both $256$.

For the adversary $D$, the CNN consists of two convolution$+$pooling layers, one MLP layer and one softmax layer, with $3\times 3$ convolution window size, $2\times 2$ pooling window size,  $20$ feature map size and $20$ MLP hidden layer size.

For the training of NMT model $G$, similar as what is commonly done in previous works~\cite{RL4NMT,NMTReconstruction}, we warm start $G$ from a well-trained RNNSearch model, and optimize it using vanilla SGD with mini-batch size $80$ for En$\rightarrow$Fr translation and $32$ for De$\rightarrow$En translation. Gradient clipping is used with clipping value 1 for En$\rightarrow$Fr and 10 for De$\rightarrow$En. The initial learning rate is chosen from cross-validation on dev set ($0.02$ for En$\rightarrow$Fr and $0.001$ for De$\rightarrow$En) and we halve it every $80k$ iterations.

An important factor we find in successfully training $G$ is that the combination of adversarial objective with MLE. That is, we force $50\%$ randomly chosen mini-batch data are trained with Adversarial-NMT, while apply MLE principle to the other mini-batches. Acting in this way significantly improves stability in model training, which is also reported in other tasks such as language model~\cite{professorForcing} and neural dialogue generation~\cite{GAN4Bot}. We conjecture that the reason is that MLE acts as a regularizer to guarantee smooth model update, alleviating the negative effects brought by high gradient estimation variance of the one-step Monte-Carlo sample in REINFORCE.

As the first step, the CNN adversary network $D$ is initially pre-trained using the sampled data $(x,y')$ sampled from the RNNSearch model, and the ground-truth translation $(x,y)$.  After that, in joint G-D training of Adversarial-NMT, the adversary is optimized using Nesterov SGD \cite{nesterov} with batch size set as $32$. The initial learning rate is $0.002$ for En$\rightarrow$Fr and $0.001$ for De$\rightarrow$En, both chosen by validation on dev set. The dimension of word embedding is the same with that of $G$, and we fix the word embeddings during training. Batch normalization~\cite{BN} is observed to significantly improve $D$'s performance. Considering efficiency, all the negative training data instances $(x,y')$ used in $D$'s training are generated using beam search with beam size $4$.

In generating model translation for evaluation, we set beam width as $4$ and $12$ for En$\rightarrow$Fr and De$\rightarrow$En respectively according to BLEU on dev set. The translation quality is measured by tokenized case-sensitive BLEU~\cite{BLEU} score \footnote{\url{https://github.com/moses-smt/mosesdecoder/blob/master/scripts/generic/multi-bleu.perl}}.

\begin{figure*}
\begin{subfigure}{.5\textwidth}
  \centering
  \includegraphics[width=1\linewidth]{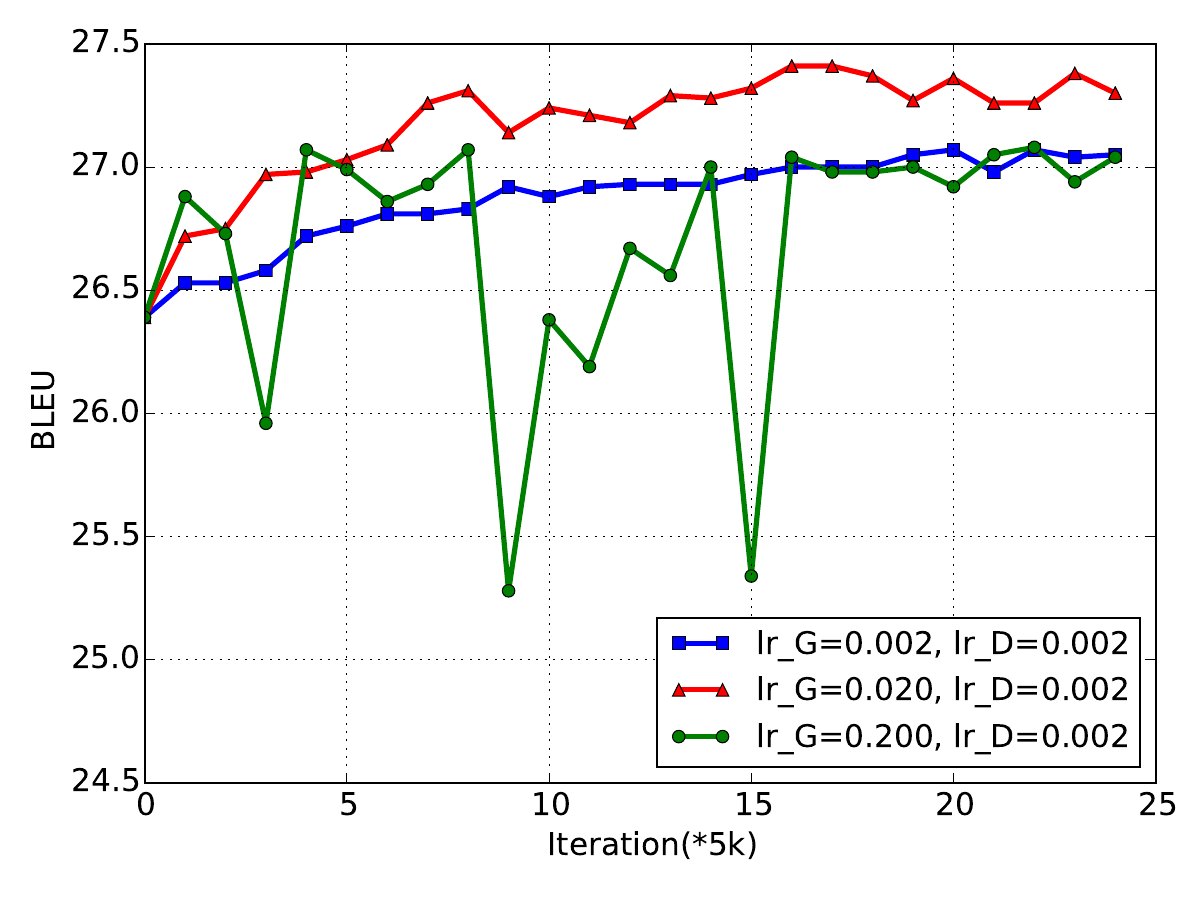}
  \caption{$D$: Same learning rates; $G$: different learning rates.}
  \label{fig:ana_fixD}

\end{subfigure}%
\begin{subfigure}{.5\textwidth}
  \centering
  \includegraphics[width=1\linewidth]{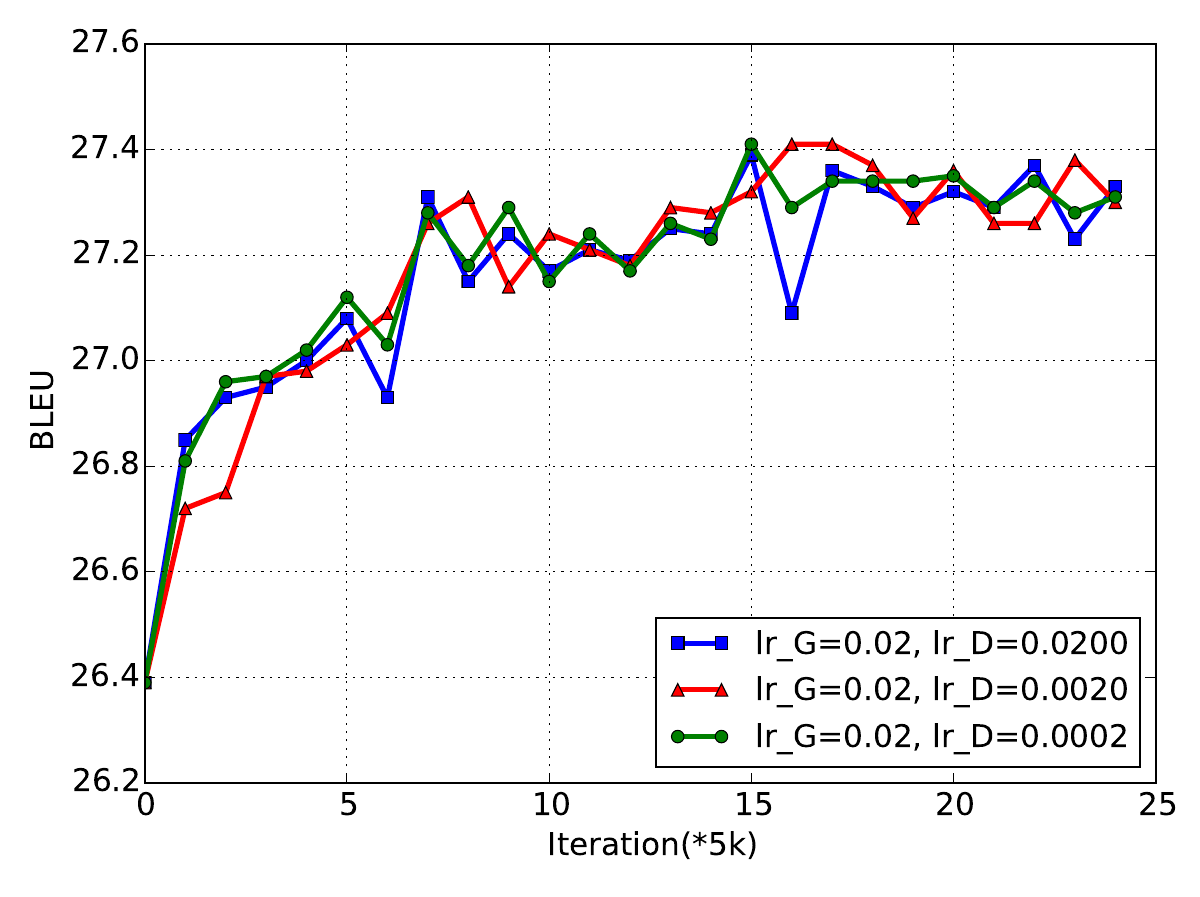}
  \caption{$G$: Same learning rates; $D$: different learning rates.}
  \label{fig:ana_fixG}
\end{subfigure}

\caption{Dev set BLEUs during En$\rightarrow$Fr Adversarial-NMT training process, with same learning rates for $D$, different learning rates for $G$ in left \ref{fig:ana_fixD}, and same learning rates for $G$ and different learning rates for $D$ in right \ref{fig:ana_fixG}.}
\label{fig:ana_lr}
\end{figure*}

\subsection{Result on En$\rightarrow$Fr translation}
\label{subsec:main_bleu}

In Table \ref{tbl:result_e2f} we provide the En$\rightarrow$Fr translation result of Adversarial-NMT, together with several strong NMT baselines, such as the well representative attention-based NMT model RNNSearch~\cite{NMT}. In addition, to make our comparison comprehensive, we would like to cover several well acknowledged techniques whose effectiveness has been verified to improve En$\rightarrow$Fr translation by previously published works, including the leverage of 1) Using large vocabulary to handle rare words ~\cite{NMTLarge,RareWordsNMT}; 2) Different training objectives~\cite{RL4NMT,PG4Sequence,AC4SequencePrediction} such as Minimum Risk Training (MRT) to directly optimize evaluation measure~\cite{RL4NMT}, and dual learning to enhance both primal and dual tasks (e.g., En$\rightarrow$Fr and Fr$\rightarrow$En)~\cite{DualNMT}; 3) Improved inference process such as beam search optimization~\cite{BSO} and postprocessing UNK~\cite{RareWordsNMT,NMTLarge}; 4) Leveraging additional monolingual data~\cite{NMTMolingual,SrcMonoNMT,DualNMT}.

From the table, we can clearly observe that Adversarial-NMT obtains satisfactory translation quality against baseline systems. In particular, it even surpasses the performances of other models with much larger vocabularies~\cite{NMTLarge}, deeper layers ~\cite{RareWordsNMT}, much larger monolingual training corpus~\cite{NMTMolingual}, and the goal of directly maximizing BLEU~\cite{RL4NMT}. In fact, as far as we know, Adversarial-NMT achieves state-of-the-art result ($34.78$) on En$\rightarrow$Fr translation for single-layer GRU sequence-to-sequence models trained with only supervised bilingual corpus on news-test 2014 test set.

{\bf Human Evaluation}: Apart from the comparison based on the objective BLEU scores, to  better appraise the performance of our model, we also involve human judgements as a subjective measure.  To be more specific, we generate the translation results for 500 randomly selected English sentences from En$\rightarrow$Fr news-test 2014 dataset using both MRT \cite{RL4NMT} and our Adversarial-NMT.  Here MRT is chosen since it is the well representative of previous NMT methods which maximize sequence level objectives, achieving satisfactory results among all single layer models (i.e., $31.3$ in Table~\ref{tbl:result_e2f}).  Afterwards we ask three human labelers to  choose the better one from the two versions of translated sentences. The evaluation process is conducted on Amazon Mechanical Turk \footnote{\url{https://www.mturk.com}}  with all the workers to be native English or French speakers.

\begin{table}[!t]
\centering 
\begin{tabular}{ c || c | c } 
\hline 
 &  Adversarial-NMT & MRT \\[0.25ex] 
\hline\hline 
evaluator 1 & 286 (57.2\%) & 214 (42.8\%) \\
evaluator 2 & 310 (62.0\%) & 190 (38.0\%) \\  
evaluator 3 & 295 (59.0\%) & 205 (41.0\%) \\
\hline
Overall & \textbf{891 (59.4\%)} & \textbf{609 (40.6\%)}\\ 
\hline 
\end{tabular}
\caption{Human evaluations for Adversarial-NMT and MRT on English$\rightarrow$French translation. ``286 (57.2\%)" means that evaluator 1 made a decision that 286 (57.2\%) out of 500 translations generated by Adversarial-NMT were better than MRT. }
\label{human_eval} 
\end{table}

Result in Table \ref{human_eval} shows that $59.4\%$ sentences are better translated by our Adversarial-NMT, compared with MRT \cite{RL4NMT}. Such human evaluation further demonstrates the effectiveness of our model and matches the expectation that Adversarial-NMT provides more human desired translation.

\begin{table*}[!t]
\small
\centering
\begin{minipage}{14cm}
\begin{tabular}{|l | l | c | c | c}
\hline
System & System Configurations  & BLEU \\
\hline \hline
\multicolumn{4}{c}{{\em Representative end-to-end NMT systems}} \\
\hline
Bahdanau et al. \shortcite{NMT} & RNNSearch & 23.87 \footnote {Reported in \protect~\cite{BSO}.} \\
Ranzato et al. \shortcite{PG4Sequence}& CNN encoder + Sequence level objective & 21.83\\
Bahdanau et al. \shortcite{AC4SequencePrediction} & CNN encoder + Sequence level actor-critic objective& 22.45\\
Wiseman et al. \shortcite{BSO}  & RNNSearch + Beam search optimization & 25.48\\
Shen et al. \shortcite{RL4NMT} &RNNSearch + Minimum Risk Training Objective& 25.84 \footnote {Result from our implementation, and we reproduced their reported En$\rightarrow$Fr result.} \\
\hline
\multicolumn{4}{c}{{\em Adversarial-NMT}} \\
\hline
\multirow{3}{* {\em this work}}{} & RNNSearch + Adversarial Training Objective & 26.98$\dagger$ \\
& RNNSearch + Adversarial Training Objective + UNK Replace & 27.94 \\
\hline
\end{tabular}
\end{minipage}
\caption{Different NMT systems' performances on De$\rightarrow$En translation. The default setting is single layer GRU encoder-decoder model with MLE training objective, i.e., the RNNSearch model proposed by Bahdanau et al. \protect\shortcite{NMT}. $\dagger$: significantly better than \protect\newcite{RL4NMT} ($\rho <$ 0.05).} \label{tbl:result_d2e}
\end{table*}

{\bf Adversarial Training: Slow or Fast}: In this subsection we analyze how to set the pace for training the NMT model $G$ and adversary $D$, to make them combatting effectively.  Specifically, for En$\rightarrow$Fr translation, we inspect how dev set BLEU varies along adversarial training process with different initial learning rates for $G$ (shown in \ref{fig:ana_fixD}) and for $D$ (shown in \ref{fig:ana_fixG}),  conditioned on the other one fixed.

Overall speaking, these two figures show that Adversarial-NMT is much more robust with regard to the pace of $D$ making progress than that of $G$, since the three curves in \ref{fig:ana_fixG} grow in a similar pattern while curves in \ref{fig:ana_fixD} drastically differ with each other. We conjecture the reason is that in Adversarial-NMT, CNN based $D$ is powerful in classification tasks, especially when it is warm started with sampled data from RNNSearch. As a comparison, the translation model $G$ is relatively weak in providing qualified translations. Therefore, training $G$ needs carefully configurations of learning rate: small value (e.g., $0.002$) leads to slower convergence (blue line in \ref{fig:ana_fixD}), while large value (e.g., $0.2$) brings un-stability (green line in \ref{fig:ana_fixD}). The proper learning rate (e.g. $0.02$) induces $G$ to make fast, meanwhile stable progress along training.

\subsection{Result on De$\rightarrow$En translation}

In Table \ref{tbl:result_d2e} we provide the De$\rightarrow$En translation result of Adversarial-NMT, compared with some strong baselines such as RNNSearch \cite{NMT} and MRT \cite{RL4NMT}.

Again, we can see that Adversarial-NMT performs best against other models from Table {\ref{tbl:result_d2e}, achieves 27.94 BLEU scores, which is also a state-of-the-art result.

\begin{table*}[!t]
\small
\centering
\begin{tabular}{|c|l|l|l|}
\hline
Source sentence $x$ & \tabincell{l}{ich weiß , dass wir es können , und soweit es mich betrifft \\ ist das etwas ,was die welt jetzt braucht .} & \multirow{2}{*}{$D(x,y')$} & \multirow{2}{*}{BLEU} \\
\cdashline{1-2}
Groundtruth translation $y$ & \tabincell{l}{i know that we can , and as far as i \&apos;m concerned ,   \\ that \&apos;s something the world needs right now .} &                    &                    \\
\hdashline
Translation by RNNSearch $y'$ &\tabincell{l}{i know we can do it , and as far as \textbf{it \&apos;s in time ,}  \\ \textbf{what} the world needs now . } &     $0.14$               &    $27.26$                \\ \hdashline
\tabincell{c}{Translation by \\ Adversarial-NMT $y'$} &  \tabincell{l}{i know that we can , and as far as \textbf{it is to be something}  \\ \textbf{that} the world needs now . }&   $0.67$                 &  $50.28$                  \\
\hline
Source sentence $x$ & \tabincell{l}{wir müssen verhindern , dass die menschen kenntnis  erlangen \\  von dingen , vor allem dann , wenn sie wahr sind .} & \multirow{2}{*}{$D(x,y')$} & \multirow{2}{*}{BLEU} \\
\cdashline{1-2}
Groundtruth translation $y$ & \tabincell{l}{we have to prevent people from finding about things , \\ especially when they are true .} &                    &                    \\
\hdashline
Translation by RNNSearch $y'$ &\tabincell{l}{we need to prevent people who are able to know \\ \textbf{ that people have to do} , especially if they are true .  } &     $0.15$               &    $0.00$                \\ \hdashline
\tabincell{c}{Translation by \\ Adversarial-NMT $y'$} &  \tabincell{l}{we need to prevent people who are able to know \\ \textbf{ about things} , especially if they are true .  }&   $0.93$                 &  $25.45$                  \\
\hline
\end{tabular}
\caption{Cases-studies to demonstrate the translation quality improvement brought by Adversarial-NMT. We provide two De$\rightarrow$En translation examples, with the source German sentence, ground-truth English sentence, and two translation results respectively provided by RNNSearch and Adversarial-NMT. $D(x,y')$ is the probability of model translation $y'$ being ground-truth translation of $x$, calculated from the adversary $D$. BLEU is per-sentence translation bleu score for each translated sentence.}\label{tbl:en-de_cases}
\end{table*}

{\bf Effect of Adversarial Training}: To better visualize and understand the advantages of adversarial training brought by Adversarial-NMT, we show several translation cases in Table \ref{tbl:en-de_cases}. Concretely speaking, we give two German$\rightarrow$English translation examples, including the source language sentence $x$, the ground-truth translation sentence $y$, and two NMT model translation sentences, respectively out from RNNSearch and Adversarial-NMT (trained after $20$ epochs) and emphasized on their different parts by bold fonts which lead to different translation quality. For each model translation $y'$, we also list $D(x,y')$, i.e., the probability that the adversary $D$ regards $y'$ as ground-truth, in the third column, and the sentence level bleu score of $y'$ in the last column.

Since RNNSearch model acts as the warm start for training Adversarial-NMT, its translation could be viewed as the result of Adversarial-NMT at its initial phase. Therefore, from Table \ref{tbl:en-de_cases}, we can observe:
\begin{itemize}
\item With adversarial training goes on, the quality of translation sentence output by $G$ gets improved, both in terms of subjective feelings and BLEU scores as a quantitative measure.
\item Correspondingly, the translation quality growth makes the adversary $D$ deteriorated, as shown by $D$'s successful recognition of $y'$ by RNNSearch as translated from model, whereas $D$ makes mistakes in classifying $y'$ out from Adversarial-NMT as ground-truth (by human).
\end{itemize}

\section{Conclusion}
\label{sec:conc}
We in this paper propose a novel and intuitive training objective for NMT, that is to force the translation results be as similar as ground-truth translations generated by human. Such an objective is achieved via an adversarial training framework called Adversarial-NMT which complements the original NMT model with an adversary based on CNN. Adversarial-NMT adopts both new network architectures to reflect the mapping within (source, target) sentence, and an efficient policy gradient algorithm to tackle the optimization difficulty brought by the discrete nature of machine translation. The experiments on both English$\rightarrow$French and German$\rightarrow$English translation tasks clearly demonstrate the effectiveness of such adversarial training method for NMT.

As to future works, with the hope of achieving new state-of-the-art performance for NMT system, we plan to fully exploit the potential of Adversarial-NMT by combining it with other powerful methods listed in subsection~\ref{subsec:main_bleu}, such as training with large vocabulary, minimum-risk principle and deep structures. We additionally would like to emphasize and explore the feasibility of adversarial training to other text processing tasks, such as image caption, dependency parsing and sentiment classification.

\bibliographystyle{acl2012}
\bibliography{tacl_gan4nmt}

\end{document}